\documentclass{article}

\usepackage[accepted]{icml2025}

\usepackage[utf8]{inputenc}

\usepackage[T1]{fontenc}
\usepackage{times}
\usepackage{amsmath, amssymb, amsthm}
\usepackage{mathtools}
\usepackage{bbm}
\usepackage{dsfont}

\PassOptionsToPackage{table,dvipsnames}{xcolor}
\definecolor{darkpink}{RGB}{199,21,140}

\usepackage{graphicx}

\usepackage{booktabs, array}

\usepackage{listings}
\usepackage{fancyvrb}
\fvset{fontsize=\small}

\definecolor{citecolor}{RGB}{0,102,204}
\definecolor{linkcolor}{RGB}{190,105,30}
\definecolor{urlcolor}{RGB}{199,21,133}

\usepackage[colorlinks,linktoc=all]{hyperref}
\usepackage[all]{hypcap}
\hypersetup{citecolor=citecolor}
\hypersetup{linkcolor=linkcolor}
\hypersetup{urlcolor=urlcolor}
\usepackage[nameinlink,capitalise]{cleveref}
\creflabelformat{equation}{#2\textup{#1}#3}  
\crefname{section}{\S}{\S\S}

\lstdefinestyle{mystyle}{
    commentstyle=\color{OliveGreen},
    numberstyle=\tiny\color{black!60},
    stringstyle=\color{BrickRed},
    basicstyle=\ttfamily\scriptsize,
    breakatwhitespace=false,
    breaklines=true,
    captionpos=b,
    keepspaces=true,
    numbers=none,
    numbersep=5pt,
    showspaces=false,
    showstringspaces=false,
    showtabs=false,
    tabsize=2
}
\theoremstyle{plain}

\theoremstyle{definition}

\theoremstyle{remark}

\def\[#1\]{\begin{equation}\begin{aligned}#1\end{aligned}\end{equation}}

\newsavebox\CBox

\usepackage[
    natbib=true,
    style=authoryear,
    backref=true,
    maxcitenames=2,
    uniquelist=false,
    sorting=nyt
]{biblatex}
\DeclareFieldFormat{citehyperref}{%
  \DeclareFieldAlias{bibhyperref}{noformat}
  \bibhyperref{#1}}

\DeclareFieldFormat{textcitehyperref}{%
  \DeclareFieldAlias{bibhyperref}{noformat}
  \bibhyperref{%
    #1%
    \ifbool{cbx:parens}
      {\bibcloseparen\global\boolfalse{cbx:parens}}
      {}}}

\savebibmacro{cite}
\savebibmacro{textcite}

\renewbibmacro*{cite}{%
  \printtext[citehyperref]{%
    \restorebibmacro{cite}%
    \usebibmacro{cite}}}

\renewbibmacro*{textcite}{%
  \ifboolexpr{
    ( not test {\iffieldundef{prenote}} and
      test {\ifnumequal{\value{citecount}}{1}} )
    or
    ( not test {\iffieldundef{postnote}} and
      test {\ifnumequal{\value{citecount}}{\value{citetotal}}} )
  }
    {\DeclareFieldAlias{textcitehyperref}{noformat}}
    {}%
  \printtext[textcitehyperref]{%
    \restorebibmacro{textcite}%
    \usebibmacro{textcite}}}

\newcommand{\neuripsbooktitle}[1]{%
  \ifnum#1<2018%
    Advances in Neural Information Processing Systems \the\numexpr#1-1987\relax\ (NIPS #1)%
  \else%
    Advances in Neural Information Processing Systems \the\numexpr#1-1987\relax\ (NeurIPS #1)%
  \fi%
}

\newcommand{\ordinalsuffix}[1]{%
  \ifcase#1 th
  \or st
  \or nd
  \or rd
  \else th
  \fi
}

\newcommand{\icmlbooktitle}[1]{%
  Proceedings of The \the\numexpr#1-1983\relax%
  \ordinalsuffix{\the\numexpr(#1-1983)\mod 10\relax} %
  International Conference on Machine Learning (ICML #1)%
}

\newcommand{\aistatsbooktitle}[1]{%
  Proceedings of The \the\numexpr#1-1997\relax%
  \ordinalsuffix{\the\numexpr(#1-1997)\mod 10\relax} %
  International Conference on Artificial Intelligence and Statistics (AISTATS #1)%
}

\newcommand{\iclrbooktitle}[1]{%
  Proceedings of The \the\numexpr#1-2012\relax%
  \ordinalsuffix{\the\numexpr(#1-2012)\mod 10\relax} %
  International Conference on Learning Representations (ICLR #1)%
}

\usepackage{wrapfig}
\usepackage{bbm}
\usepackage{multirow}
\usepackage{makecell}
\usepackage{pifont} 
\usepackage{colortbl}

\usepackage[most]{tcolorbox}
\usepackage{graphicx}
\usepackage{lipsum}

\usepackage{listings}
\tcbuselibrary{listings}

\tcbset{
  highlight token/.style={
    boxrule=0.4pt,
    colback=gray!5,
    colframe=gray!50,
    fontupper=\ttfamily\color{purple},
    arc=1pt,
    boxsep=1pt,
    left=0.5pt,
    right=0.5pt,
    top=0.5pt,
    bottom=0.5pt,
    enhanced,
    box align=base,
    on line,
    height=1.2em,
    valign=center, 
  }
}

\newcommand{\smallslash}{{\fontsize{7pt}{7pt}\selectfont /}}

\definecolor{promptbg}{RGB}{245,245,245}
\definecolor{promptborder}{RGB}{200,200,200}
\definecolor{prompttitle}{RGB}{40,40,40}
\definecolor{prompttext}{RGB}{20,20,20}

\newtcolorbox[auto counter]{prompt}[2][]{
  listing only,
  listing options={
    basicstyle=\ttfamily\footnotesize\color{prompttext},
    breaklines=true,
    showstringspaces=false,
  },
  colback=promptbg,
  colframe=promptborder,
  coltitle=prompttitle,
  title=>>>~#2,
  fonttitle=\bfseries,
  boxrule=0.4pt,
  arc=2pt,
  left=1em,
  right=1em,
  top=0.7em,
  bottom=0.7em,
  #1,
}

\icmltitlerunning{}

\begin{document}

\twocolumn[
\icmltitle{\texorpdfstring{Verbalized Confidence Triggers Self-Verification
\\: Emergent Behavior Without Explicit Reasoning Supervision}{Verbalized Confidence Triggers Self-Verification Emergent Behavior Without Explicit Reasoning Supervision}}

\begin{icmlauthorlist}
\icmlauthor{Chaeyun Jang}{yyy}
\icmlauthor{Moonseok Choi}{yyy}
\icmlauthor{Yegon Kim}{yyy}
\icmlauthor{Hyungi Lee}{yyy}
\icmlauthor{Juho Lee}{yyy}
\end{icmlauthorlist}

\icmlaffiliation{yyy}{
    Korea Advanced Institute of Science and Technology, Daejeon, Korea}

\icmlcorrespondingauthor{Juho Lee}{juholee@kaist.ac.kr}

\icmlkeywords{Machine Learning, ICML}

\vskip 0.3in
]

\printAffiliationsAndNotice{}




\begin{abstract}
Uncertainty calibration is essential for the safe deployment of large language models (LLMs), particularly when users rely on verbalized confidence estimates. While prior work has focused on classifiers or short-form generation, confidence calibration for chain-of-thought (CoT) reasoning remains largely unexplored. Surprisingly, we find that supervised fine-tuning with scalar confidence labels alone suffices to elicit self-verification behavior of language models, without any explicit reasoning supervision or reinforcement learning-based rewards. Despite being trained only to produce a verbalized confidence score without any self-verifying examples, the model learns to generate longer and self-checking responses for low-confidence queries while providing more concise answers for high-confidence ones. We further propose a simple rethinking method that boosts performance via test-time scaling based on calibrated uncertainty. Experiments on GSM8K and held-out reasoning tasks such as MATH-500 and ARC-Challenge show that our confidence-aware fine-tuning improves both calibration and accuracy, while also enhancing interpretability by aligning the model's reasoning path with its confidence. 
\end{abstract}

\section{Introduction}

Large language models (LLMs) demonstrate strong performance not only in natural language generation but also in complex reasoning and decision-support tasks across diverse domains~\citep{achiam2023gpt, guo2025deepseek}. Their application to high-stakes settings such as medical diagnosis and personalized financial analysis has drawn increasing attention~\citep{goh2024large, qiu2024llm, takayanagi2025generative}, with the potential to reduce expert workload and accelerate decision-making.

Nonetheless, ensuring the reliability of LLMs remains a critical challenge. Models frequently produce incorrect outputs with high confidence, and such overconfident errors can lead to harmful decisions when left undetected by users~\citep{zhou2024relying}. These hallucinations pose risks that go beyond factual inaccuracies, with implications for healthcare, law, and finance~\citep{du2024haloscope}. To mitigate these risks, LLMs must be able to quantify and communicate their uncertainty in a human interpretable manner, such as through \emph{verbalized confidence}.

\begin{figure}[t]
\centering
\includegraphics[width=0.8\linewidth]{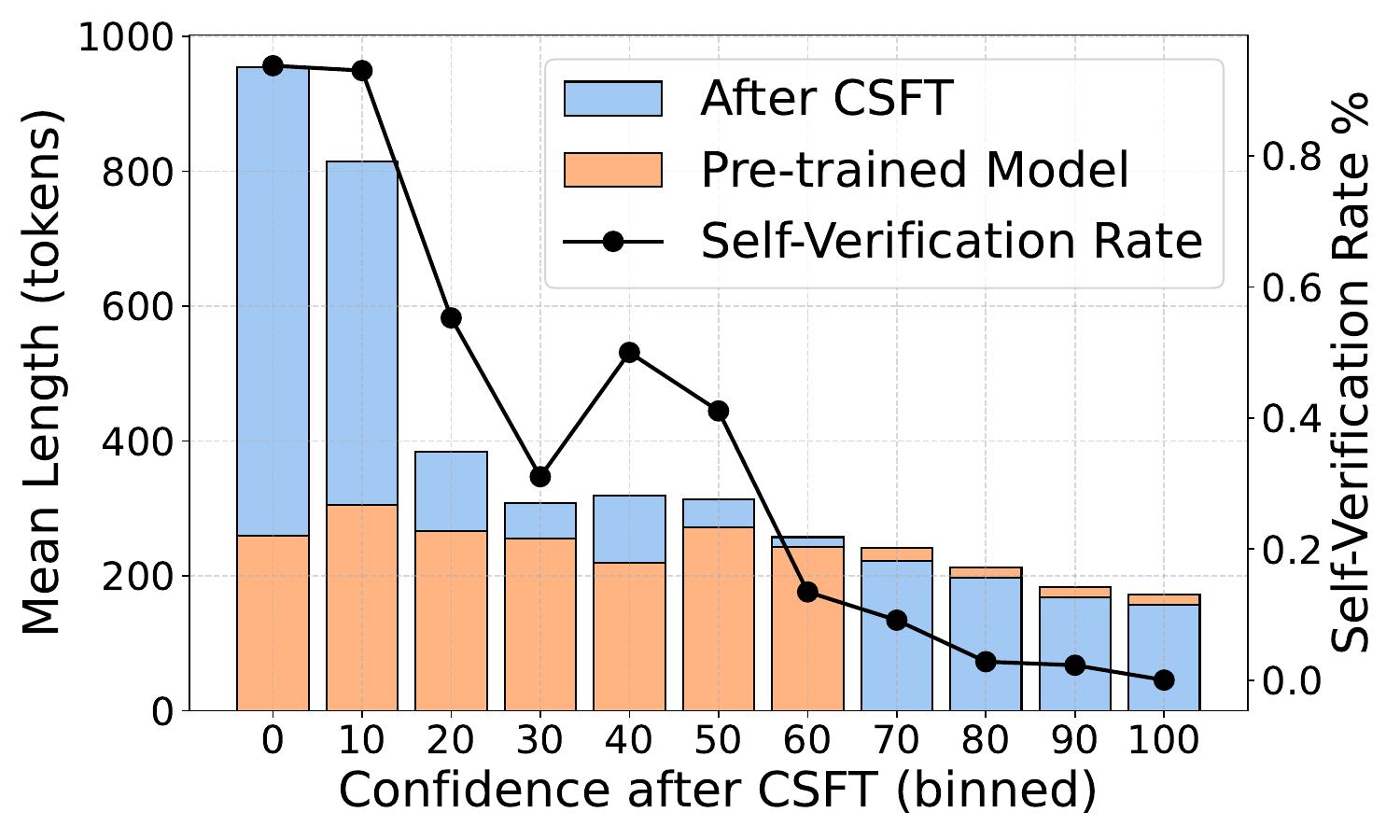}
\vspace{-0.5em}
\caption{
Generation length and self-verification rate across confidence bins on GSM8K using the CSFT-trained \texttt{LLaMA-3.2-3B-Instruct} model. Lower-confidence bins yield longer outputs and higher self-verification rates, suggesting a learned fallback behavior. Overall, 20\% of generations showed self-verification (measured using \texttt{GPT-4.1}), compared to under 1.5\% in the zero-shot setting, which is omitted. Representative examples are shown in \Cref{fig:examples}.
}
\label{fig:intro_length}
\end{figure}

While a few recent studies train models to explicitly verbalize their confidence~\citep{band2024linguistic, kapoorlarge, stengel2024lacie, jang2024calibrated}, most rely on complex procedures such as reinforcement learning or classifier probing. Moreover, they report little evidence of generalization in zero-shot settings or under chain-of-thought (CoT) reasoning. At the same time, state-of-the-art LLMs increasingly tackle challenging problems by generating explicit CoT traces~\citep{zhou2023leasttomost, chen2023progofthought, wang2023selfconsistency}. The reliability of verbalized confidence within such CoT reasoning, however, remains largely unexplored.

In contrast, this work shows that even a simple confidence-supervised fine-tuning (CSFT) approach, under suitable conditions, can yield well-calibrated verbalized confidence in CoT reasoning. Furthermore, we demonstrate that the model can autonomously adjust its response length and exhibit emergent self-verification behavior as a function of its uncertainty. Specifically, without reasoning supervision, the model learns to modulate its reasoning process while being trained to generate an answer, followed by a verbalized confidence score. As illustrated in \Cref{fig:intro_length}, low-confidence predictions result in longer outputs with self-check phrases such as ``recalculate'' or ``let me double-check'', whereas high-confidence responses are shorter and more decisive. This phenomenon emerges even without CoT reasoning guidance in training, and is consistently observed across GSM8K, MATH-500, and ARC-Challenge. These findings suggest that verbalized confidence can serve not only as a calibration target but also as an effective training signal that encourages more deliberate chain-of-thought generation, ultimately leading to improved reasoning accuracy.

Our contributions are as follows:
\begin{itemize}
\item We propose CSFT, a simple confidence-supervised fine-tuning method using the problems and the corresponding synthetic self-confidence labels that enables reliable verbalized confidence in CoT reasoning tasks.
\item We demonstrate that CSFT elicits emergent self-verification behavior without requiring reasoning supervision, manifesting as a systematic relationship between confidence and output length.
\item We analyze how prompting style, regularization strength, and reasoning depth affect this phenomenon, and demonstrate generalization to held-out reasoning tasks.
\end{itemize}

These results provide a scalable path toward building uncertainty-aware LLMs using standard SFT pipelines, without architectural modifications or post-hoc correction. CSFT not only improves calibration but also guides model behavior toward safer and more interpretable reasoning.
\section{Related Works}

\subsection{Confidence Calibration in LLMs}

Calibration in LLMs has been studied from various perspectives. Likelihood-based methods estimate uncertainty using token-level entropy, sequence probabilities, or generation variance~\citep{desai2020calibration, nguyen2024semantic, kadavath2022language}. These methods are helpful for model-side diagnostics, but they do not yield human-readable confidence statements.
Verbalized confidence, where models explicitly articulate how sure they are, has emerged as a more user-friendly and interpretable alternative.~\citep{band2024linguistic, stengel2024lacie}. However, most existing approaches focus on short-form declarative QA and require manual scalar labels~\citep{lin2024confidence} or classifier-based tuning~\citep{kapoorlarge, jang2024calibrated}, without an understanding of the dynamics of reasoning.
In the context of CoT reasoning, recent work has reported that instruction-tuned or reasoning-supervised models~\citep{yang2025qwen3, guo2025deepseek} exhibit better calibration under zero-shot inference~\citep{yoon2025reasoning}. However, these findings remain observational and do not examine how calibration can be systematically induced or controlled in reasoning tasks.
In contrast, our work provides the first direct evidence that fine-tuning with weak, self-derived confidence labels, obtained through consistency across sampled answers, can induce improved calibration and emergent self-verification in reasoning, even in models with no prior exposure to CoT supervision.

\subsubsection{Self-Verification and CoT Optimization}
Self-verification has emerged as a desirable property for LLMs, with prior work demonstrating that models capable of revisiting and refining their reasoning tend to achieve higher accuracy and robustness, especially on complex tasks~\citep{wang2023selfconsistency}. 
Accordingly, a growing body of research has focused on optimizing the structure of CoT outputs, either by making them more concise~\citep{nayab2024concise, team2025kimi} or by generating longer and more reflective traces~\citep{guo2025deepseek}. 
However, these approaches typically target a fixed generation style, without conditioning on the model’s internal uncertainty.
In reality, effective reasoning should adapt to the model's confidence: when confidence is low, the model should elaborate and verify; when confidence is high, a brief and decisive answer may suffice.
Most existing methods do not capture this dynamic.
Reinforcement learning-based approaches train models to favor verifiable traces~\citep{chen2023progofthought, zhao2025learning, shafayat2025can}, but require reward shaping and large-scale tuning.
We show that a single round of fine-tuning on scalar confidence labels derived from self-consistency induces CoT behaviors that adaptively reflect the model’s uncertainty.
\section{Confidence-Supervised Fine-Tuning}
\label{method}

We introduce Confidence-Supervised Fine-Tuning (CSFT), a simple yet effective method for calibrating verbalized confidence in LLMs under reasoning scenarios, without requiring explicit supervision of the reasoning process. CSFT fine-tunes the model to calibrate its verbalized confidence, while also producing a CoT reasoning trace and final answer, so that the reported confidence more accurately reflects the model's belief in the correctness of its answer.

Given a question $q$, the decoder is trained to generate a structured response consisting of: (i) a CoT reasoning trace $r$ and final answer $a$, enclosed within {\small\tcbox[highlight token]{<think>}}~...~{\small\tcbox[highlight token]{<\smallslash think>}} and {\small\tcbox[highlight token]{<answer>}}~...~{\small\tcbox[highlight token]{<\smallslash answer>}} tags, respectively; followed by (ii) a suffix confidence prompt (see \Cref{app:prompts}), which elicits a discrete confidence score $c \in \{0, 10, \dots, 100\}$ expressed in {\small\tcbox[highlight token]{<confidence>}}~...~{\small\tcbox[highlight token]{<\smallslash confidence>}} tags. Only the confidence score $c$ is supervised during training; both the reasoning $r$ and the answer $a$ remain unconstrained. 

In this section, we used Low Rank Adaptation~\citep[LoRA;][]{hu2021lora} method to fine-tune the LLM model $f_{\theta_0}$. Here, we denote $\theta_0$ as fixed pre-trained parameters and $\theta$ as fixed pre-trained parameters with additional learnable LoRA weights.

\paragraph{Self-Confidence label.}
To compute the confidence label, we first sample $K$ full generations $\{(r^{(i)}, a^{(i)})\}_{i=1}^{K} \sim f_\theta(\cdot \mid q)$ and estimate the empirical accuracy as
\[
\hat{p}(q) =
\frac{1}{K} \sum_{i=1}^{K}
\mathbbm{1}\bigl[a^{(i)} = a^\star\bigr],
\]
where $a^\star$ denotes the gold answer. To determine whether each answer $a^{(i)}$ matches the gold answer $a^\star$, we first parsed the value between {\small\tcbox[highlight token]{<answer>}}~...~{\small\tcbox[highlight token]{</answer>}} from the LLM response to extract $a^{(i)}$, and then checked whether it exactly matched the gold answer.
Then, the self-confidence label was obtained by discretizing the accuracy: $c = \lfloor 100 \cdot \hat{p}(q) \rfloor$.

\paragraph{Training objective.}
Let $T_c$ denote the token positions corresponding to the entire confidence span, including the {\small\tcbox[highlight token]{<confidence>}}~...~{\small\tcbox[highlight token]{<\smallslash confidence>}} tags.  
CSFT minimizes the masked cross-entropy loss over these positions:
\[
\mathcal{L}_{\text{CSFT}} =
- \sum_{t \in T_c} \log p_\theta(y_t \mid y_{<t}, q),
\]
where $p_\theta(y)$ indicates the predicted probability of token $y$ from LLM $f_\theta$. And optionally, we add a KL regularization term over the CoT and answer spans (including their respective tags), where the corresponding token positions are denoted by $T_{\text{KL}}$, to encourage the model to remain close to the pretrained distribution:
\[
\mathcal{L}_{\text{total}} =
\mathcal{L}_{\text{CSFT}} +
\lambda \sum_{t \in T_{\text{KL}}}
\mathrm{KL}\!\left(
p_\theta
\;\|\;
p_{\theta_0}
\right),
\]
where $\lambda$ is a weighting hyperparameter, and $p_\theta$ and $p_{\theta_0}$ represents $p_\theta(\cdot \mid y_{<t}, q)$ and $p_{\theta_0}(\cdot \mid y_{<t}, q)$, respectively. Unless otherwise noted, we set $\lambda = 0$.

\paragraph{Calibration effect.}
Because the target $c$ reflects the empirical accuracy of the model’s own generations, CSFT encourages alignment between predicted confidence and actual correctness. This leads to improved calibration, as measured by standard metrics such as expected calibration error~\citep[ECE;][]{ece} in CoT reasoning tasks.
\section{Experiments}

\paragraph{Experimental Setup.} 

We construct our training corpus by sampling $K=10$ CoT traces and their corresponding answers for each question $q$ from the GSM8K~\citep{cobbe2021training} training split, and assigning a self-consistency label based on the proportion of sampled answers that match the gold answer. Using this signal, we fine-tune \texttt{LLaMA3.2-3B-Instruct}~\citep{grattafiori2024llama} and \texttt{Qwen2.5-1.5B-Instruct}~\citep{yang2025qwen3}, and evaluate accuracy, Area Under the Receiver Operating Characteristic curve (AUROC), and calibration metrics (ECE, Brier Score) on the GSM8K test set as well as on the held-out reasoning benchmarks MATH-500~\citep{lightman2023let} and ARC-Challenge~\citep{clark2018think}. Further experimental details are provided in \Cref{app:experimental_details}.

\subsection{Main Experiments}

\begin{table*}[t]
\centering
\small                
\setlength{\tabcolsep}{5pt}
\caption{Calibration results on the in-distribution GSM8K and two held-out reasoning benchmarks. For datasets, \ding{51}: `seen' during CSFT, \ding{55}: `unseen' during CSFT. For the metrics, $\downarrow$: lower is better and $\uparrow$: higher is better.}
\vspace{0.5em}
\label{tab:main_results}
\begin{tabular*}{\textwidth}{@{\extracolsep{\fill}} ll lccccc}
\toprule
\textbf{Dataset} & \textbf{Model} & \textbf{Method} & \textbf{AUROC}\,($\uparrow$) & \textbf{ACC}\,($\uparrow)$ & \textbf{ECE}\,($\downarrow$) & \textbf{BS}\,($\downarrow$) & \textbf{Avg. Len.} \\
\midrule

\multirow{4}{*}[0.5ex]{\textbf{GSM8K} (\ding{51})} 
  & \multirow{2}{*}[0.5ex]{\texttt{LLaMA3.2-3B-Instruct}} & Pre-trained & 50.57 & 68.68 & 0.2065 & 0.2549 & 226.66 \\
  &                                              & CSFT      & 81.25 & 71.34 & 0.0568 & 0.1450 & 288.71 \\
\cmidrule(lr){2-8}
  & \multirow{2}{*}[0.5ex]{\texttt{Qwen2.5-1.5B-Instruct}}     & Pre-trained & 49.59 & 67.85 & 0.1928 & 0.2915 & 250.21 \\
  &                                              & CSFT      & 67.67 & 69.63 & 0.0552 & 0.2285 & 291.70 \\

\midrule

\multirow{4}{*}[0.5ex]{\textbf{MATH-500} (\ding{55})} 
  & \multirow{2}{*}[0.5ex]{\texttt{LLaMA3.2-3B-Instruct}} & Pre-trained & 49.57 & 41.20 & 0.4730 & 0.4800 & 416.70 \\
  &                                              & CSFT       & 62.97 & 56.60 & 0.1776 & 0.3059 & 559.33 \\
\cmidrule(lr){2-8}
  & \multirow{2}{*}[0.5ex]{\texttt{Qwen2.5-1.5B-Instruct}}     & Pre-trained & 59.91 & 55.00 & 0.3786 & 0.2978 & 444.86 \\
  &                                              & CSFT       & 60.27 & 56.40 & 0.2590 & 0.2629 & 477.96 \\

\midrule

\multirow{4}{*}[0.5ex]{\textbf{ARC-Challenge} (\ding{55})} 
  & \multirow{2}{*}[0.5ex]{\texttt{LLaMA3.2-3B-Instruct}} & Pre-trained & 53.89 & 65.36 & 0.2251 & 0.2738 & 210.50 \\
  &                                              & CSFT      & 72.58 & 69.45 & 0.0647 & 0.1853 & 293.13 \\
\cmidrule(lr){2-8}
  & \multirow{2}{*}[0.5ex]{\texttt{Qwen2.5-1.5B-Instruct}}     & Pre-trained & 54.08 & 52.07 & 0.1660 & 0.2680 & 105.55 \\
  &                                              & CSFT      & 61.63 & 56.82 & 0.1107 & 0.2584 & 114.85 \\

\bottomrule
\end{tabular*}
\end{table*}

\paragraph{Evaluation on GSM8K dataset.}
\Cref{tab:main_results} show results on the GSM8K test set, where our model was trained using CSFT with the GSM8K training dataset. Our method consistently improves all calibration metrics and accuracy over the pre-trained baseline, not only on \texttt{LLaMA3.2-3B-Instruct} but also on \texttt{Qwen2.5-1.5B-Instruct}. Beyond better alignment between predicted confidence and correctness, we observe that CSFT induces desirable reasoning behaviors, such as internal error checking and more deliberate output construction—hallmarks of self-verification, particularly in low-confidence cases. Refer to \cref{fig:examples} to see an example. And in such low-confidence cases, responses based on self-verification can serve as a mechanism for users to view the predicted model confidence as more reliable.

\paragraph{Unseen CoT Tasks.} 
To evaluate the generalization capability of the model trained with CSFT, we test it on two reasoning benchmarks—MATH-500 and ARC-Challenge—that are structurally, topically, and cognitively distinct from GSM8K. These benchmarks allow us to assess how well the learned reasoning patterns transfer to the unseen domain and more challenging problem distributions.
As shown in \Cref{tab:main_results}, CSFT improves generalization to unseen CoT tasks for both models. On MATH-500, \texttt{LLaMA3.2-3B-Instruct} achieves a +37$\%$ accuracy gain and 63$\%$ ECE reduction, while \texttt{Qwen2.5-1.5B-Instruct} yields a +2.5$\%$ accuracy gain and 32$\%$ lower ECE. On ARC-Challenge, LLaMA improves accuracy by 6.3$\%$ and ECE by 71$\%$, and Qwen shows 9.1$\%$ higher accuracy and 33$\%$ better calibration. 
These results indicate that our self-consistency-based calibration transfers beyond the training task, improving both confidence alignment and problem-solving ability on diverse reasoning challenges.

\subsection{Self-Verification Behaviour} 
In this section, we analyze how training with CSFT induces self-verification behavior and how this behavior correlates with predicted confidence levels. We further support our analysis with concrete examples. First, \Cref{fig:intro_length} presents two key relationships: (1) the average CoT token length as a function of predicted confidence, and (2) the proportion of answers that trigger self-verification across different confidence levels.

The results in \Cref{fig:intro_length} show a clear trend: CSFT-trained models generate significantly longer outputs when their confidence is low. In particular, for the lowest-confidence bin (0), the average output length is nearly five times longer than that of the zero-shot baseline. This indicates that the model compensates for low confidence by engaging in extended reasoning, suggesting that self-verification is an emergent behavior tied to uncertainty. This increase in length is closely accompanied by a high self-verification rate, with nearly all corrected answers involving explicit verification behaviors. Moreover, this self-verification behavior under low-confidence scenarios can serve as a valuable mechanism for end users relying on LLM responses. It provides an implicit signal that the model is uncertain and is actively working to validate its answer, thereby enhancing the perceived reliability of the predicted confidence from the user's perspective. As shown in \Cref{app:length}, both the length and self-verification patterns generalize beyond the training distribution, emerging similarly on unseen CoT tasks as well.

\paragraph{Qualitative Examples.}

\begin{figure*}[t]
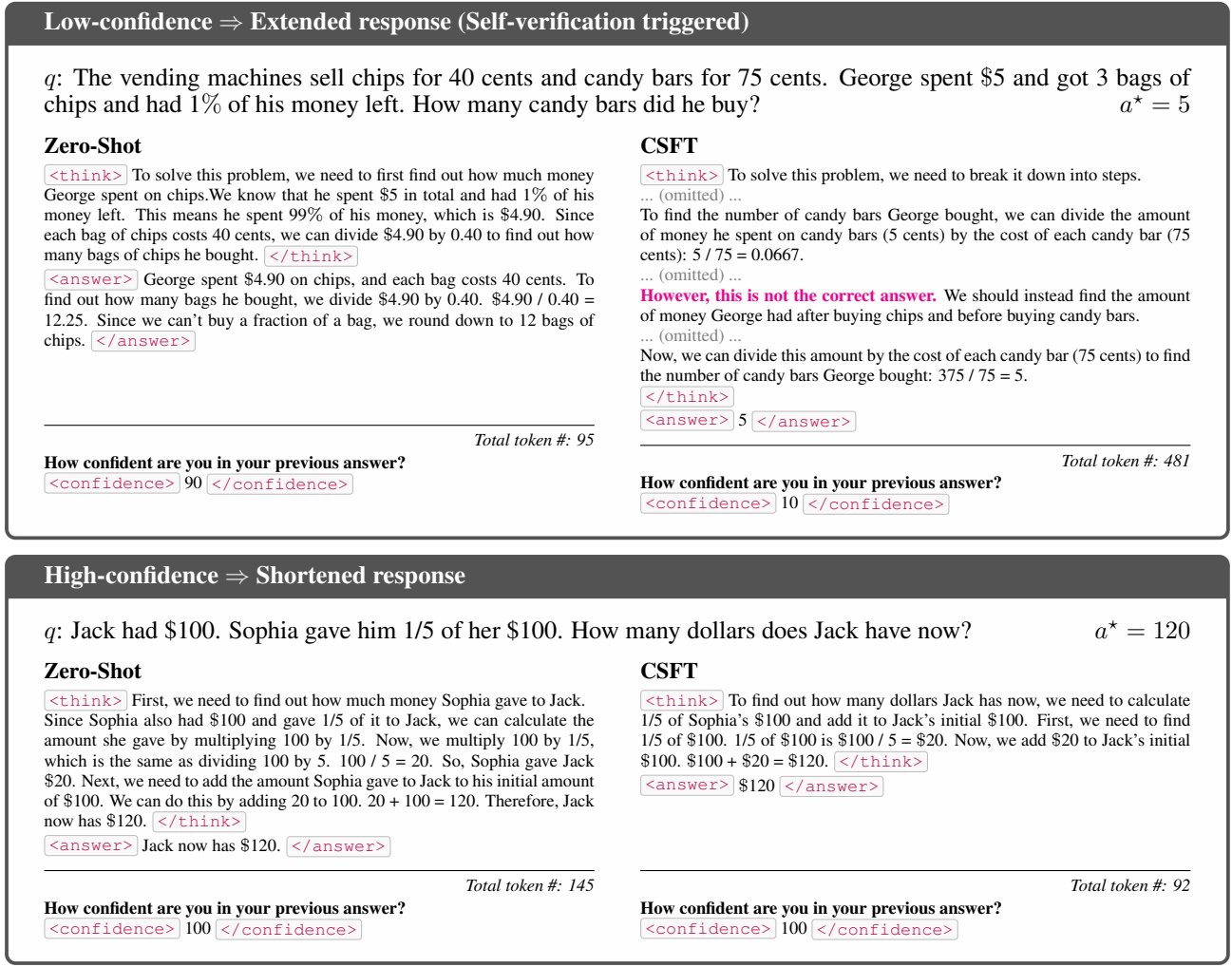

\centering
\begin{tcolorbox}[colback=gray!0.5, colframe=darkgray!90, title=Low-confidence $\Rightarrow$ Extended response (Self-verification triggered), fonttitle=\bfseries, width=\textwidth]

$q$: The vending machines sell chips for 40 cents and candy bars for 75 cents. George spent $\$$5 and got 3 bags of chips and had 1$\%$ of his money left. How many candy bars did he buy? \hfill $a^{\star}  = 5$
\\
\\
\scriptsize
\begin{minipage}[t]{0.48\textwidth}
{\small\textbf{Zero-Shot}}\par
\smallskip

\tcbox[highlight token]{<think>} To solve this problem, we need to first find out how much money George spent on chips.We know that he spent $\$$5 in total and had 1$\%$ of his money left. This means he spent 99$\%$ of his money, which is $\$$4.90. Since each bag of chips costs 40 cents, we can divide $\$$4.90 by 0.40 to find out how many bags of chips he bought.
\tcbox[highlight token]{</think>}\\
\tcbox[highlight token]{<answer>} George spent $\$$4.90 on chips, and each bag costs 40 cents. To find out how many bags he bought, we divide $\$$4.90 by 0.40. $\$$4.90 / 0.40 = 12.25. Since we can't buy a fraction of a bag, we round down to 12 bags of chips. \tcbox[highlight token]{</answer>}
\\
\\
\\
\\
\rule{\linewidth}{0.4pt}
\vspace{1pt}
\hfill\textit{Total token \#: 95}\\
\textbf{How confident are you in your previous answer?}\\
\tcbox[highlight token]{<confidence>} 90 \tcbox[highlight token]{</confidence>}
\end{minipage}
\hfill
\begin{minipage}[t]{0.48\textwidth}
{\small\textbf{CSFT}}\par
\smallskip
\tcbox[highlight token]{<think>} To solve this problem, we need to break it down into steps.
\\
\textcolor{gray}{... (omitted) ...}\\
To find the number of candy bars George bought, we can divide the amount of money he spent on candy bars (5 cents) by the cost of each candy bar (75 cents): 5 / 75 = 0.0667.\\
\textcolor{gray}{... (omitted) ...}\\
\textbf{\textcolor{magenta}{However, this is not the correct answer.}} We should instead find the amount of money George had after buying chips and before buying candy bars.\\
\textcolor{gray}{... (omitted) ...}\\
Now, we can divide this amount by the cost of each candy bar (75 cents) to find the number of candy bars George bought: 375 / 75 = 5.\\
\tcbox[highlight token]{</think>} \\
\tcbox[highlight token]{<answer>} 5 \tcbox[highlight token]{</answer>} \\
\rule{\linewidth}{0.4pt}
\vspace{1pt}
\hfill\textit{Total token \#: 481}

\textbf{How confident are you in your previous answer?}\\
\tcbox[highlight token]{<confidence>} 10 \tcbox[highlight token]{</confidence>}
\end{minipage}
\end{tcolorbox}

\begin{tcolorbox}[colback=gray!0.5, colframe=darkgray!90, title=High-confidence $\Rightarrow$ Shortened response, fonttitle=\bfseries, width=\textwidth]

$q$: Jack had \$100. Sophia gave him 1/5 of her \$100. How many dollars does Jack have now? \hfill $a^{\star}  = 120$
\\
\\
\scriptsize
\begin{minipage}[t]{0.48\textwidth}
{\small\textbf{Zero-Shot}}\par
\smallskip
\tcbox[highlight token]{<think>} First, we need to find out how much money Sophia gave to Jack. \\
Since Sophia also had $\$$100 and gave 1/5 of it to Jack, we can calculate the amount she gave by multiplying 100 by 1/5. Now, we multiply 100 by 1/5, which is the same as dividing 100 by 5. 100 / 5 = 20. So, Sophia gave Jack $\$$20. Next, we need to add the amount Sophia gave to Jack to his initial amount of $\$$100. We can do this by adding 20 to 100. 20 + 100 = 120. Therefore, Jack now has $\$$120.
\tcbox[highlight token]{</think>} \\
\tcbox[highlight token]{<answer>} Jack now has $\$$120. \tcbox[highlight token]{</answer>}\\
\rule{\linewidth}{0.4pt}
\vspace{1pt}
\hfill\textit{Total token \#: 145}\\
\textbf{How confident are you in your previous answer?}\\
\tcbox[highlight token]{<confidence>} 100 \tcbox[highlight token]{</confidence>}
\end{minipage}
\hfill
\begin{minipage}[t]{0.48\textwidth}
{\small\textbf{CSFT}}\par
\smallskip
\tcbox[highlight token]{<think>} To find out how many dollars Jack has now, we need to calculate 1/5 of Sophia's $\$$100 and add it to Jack's initial $\$$100. First, we need to find 1/5 of $\$$100. 1/5 of $\$$100 is $\$$100 / 5 = $\$$20. Now, we add $\$$20 to Jack's initial $\$$100. $\$$100 + $\$$20 = $\$$120. \tcbox[highlight token]{</think>} \\
\tcbox[highlight token]{<answer>} $\$$120 \tcbox[highlight token]{</answer>} \\
\\
\\
\\
\rule{\linewidth}{0.4pt}
\vspace{1pt}
\hfill\textit{Total token \#: 92}

\textbf{How confident are you in your previous answer?}\\
\tcbox[highlight token]{<confidence>} 100 \tcbox[highlight token]{</confidence>}
\end{minipage}
\end{tcolorbox}\caption{\textbf{Confidence reshapes reasoning behavior.} Left: In the low-confidence case (top), the CSFT model generates a long reasoning trace with explicit self-verification, eventually arriving at the correct answer, while the zero-shot model fails.
Right: In the high-confidence case (bottom), both models give the correct answer, but the CSFT response is significantly more concise, reflecting confidence-aware brevity.}
\label{fig:examples}
\end{figure*}

In \Cref{fig:examples}, we provide qualitative examples comparing responses generated by the pre-trained baseline and the model fine-tuned with CSFT. The results highlight that CSFT-trained models dynamically adjust both the length and structure of their reasoning based on the predicted confidence.
Low-confidence cases elicit longer, reflective traces with internal correction, whereas high-confidence responses tend to be brief and decisive.

\subsection{Ablation Studies}

\begin{figure}
\centering
\includegraphics[width=0.8\linewidth]{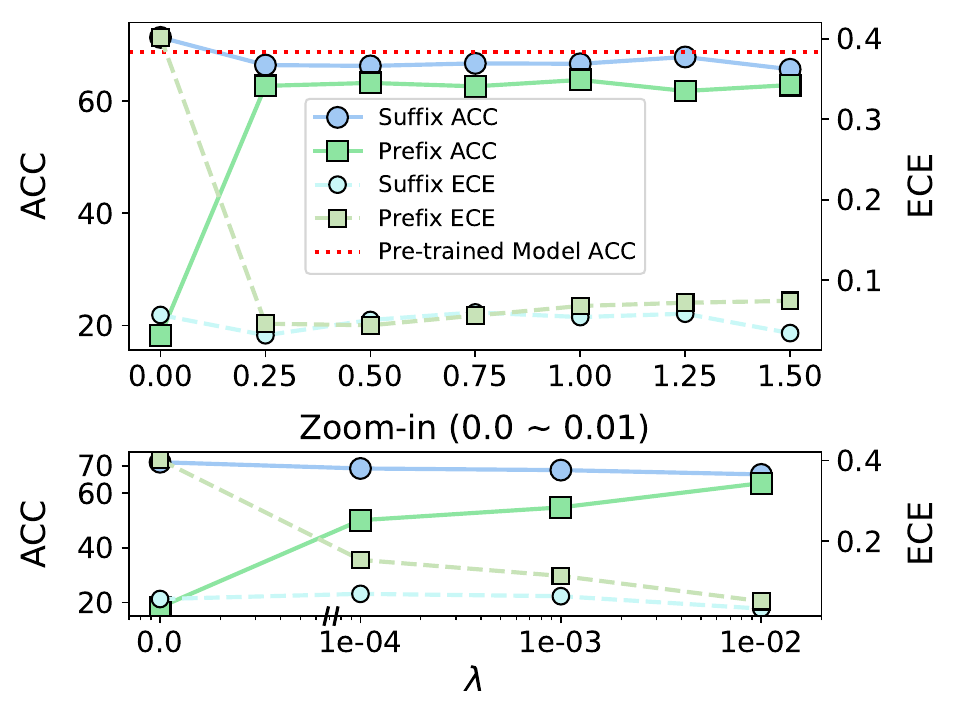}
\vspace{-0.5em}
\caption{
Test accuracy and ECE on GSM8K using CSFT-trained \texttt{LLaMA3.2-3B-Instruct}, evaluated under varying KL weights. 
Prefix performance declines without KL, whereas suffix remains stable or improves.
}
\label{fig:kl_position}
\end{figure}

\paragraph{Prompt Position.}
The \textit{suffix} setting, in which the confidence prompt is appended after the model's answer, is described in \Cref{method}. 
In addition, we evaluate a \textit{prefix} variant where the confidence prompt is inserted immediately after the question and before any reasoning begins (see \Cref{app:prompts}). 
This placement allows the model to condition its reasoning on anticipated confidence, which may affect both generation and calibration.
As shown in \Cref{fig:kl_position}, the prefix setting consistently yields lower accuracy than the zero-shot baseline, with performance sharply degrading when KL regularization is removed (i.e., $\lambda = 0$).
This suggests that without constraints on the CoT and answer spans, the model may overfit to expressing uncertainty rather than reasoning accurately, effectively learning to be confidently wrong or confidently uncertain.
In contrast, the suffix setting shows more favorable behavior. When KL regularization is removed, both calibration and accuracy improve compared to the zero-shot baseline. 
These results highlight a key trade-off: prefix prompting influences the generation process and requires regularization to remain effective, while suffix prompting is more stable because it does not interfere with the model’s reasoning.

\paragraph{KL Regularization.}
\Cref{fig:kl_position}
We investigate the effect of varying the KL regularization weight $\lambda$ on model performance, focusing on its impact near $\lambda = 0$ (see \Cref{fig:kl_position}, Zoom-in figure). 
The motivation for this analysis is to examine whether performance gains at low $\lambda$ are stable or merely an artifact of tuning.
For the prefix setting, we observe that performance rapidly deteriorates as KL regularization is removed. This suggests that, without a constraint to preserve the pretrained distribution over CoT and answer spans, the model may exploit the freedom to optimize ECE at the expense of actual reasoning quality. In effect, the model becomes well-calibrated but confidently incorrect.
In contrast, the suffix setting remains stable or even improves in the absence of KL regularization. Since confidence is predicted independently after the full generation, removing the KL constraint does not impair reasoning quality, and may in fact allow for better post hoc alignment of confidence with correctness.
These results highlight the importance of controlling model behavior when confidence supervision is introduced at generation time (prefix), as opposed to after-the-fact (suffix).

\paragraph{Label Quality and Confidence Question.}

\begin{table}[ht]
\centering
\caption{Ablation analysis on the impact of confidence label quality and the inclusion of confidence prompt. Results are reported as differences relative to CSFT on GSM8K using \texttt{LLaMA3.2-3B-Instruct}.}
\vspace{0.5em}
\label{tab:label_quality_conf}
\renewcommand{\arraystretch}{1.3}
\setlength{\tabcolsep}{5pt}
\small
\begin{tabular}{lccc}
\toprule
\textbf{Variant} & \textbf{ACC} & \textbf{ECE} & \textbf{Avg. Len.} \\
\midrule
w/o Correct label    & \textbf{-}2.14 & \textbf{+}0.05 & \textbf{-}49.36 \\
w/o Conf question     & \multicolumn{3}{c}{Training collapsed} \\
\bottomrule
\end{tabular}
\end{table}

To evaluate the role of confidence supervision and the design of the confidence prompt in enabling stable training and achieving strong performance, we conduct two ablation studies, as summarized in \Cref{tab:label_quality_conf}.
First, we randomly assign confidence labels within the {\small\tcbox[highlight token]{<confidence>}} tag, breaking the link between prediction quality and label supervision.
Second, we remove the explicit confidence prompt, instead asking the model to generate a scalar confidence directly after the final answer using a repeated {\small\tcbox[highlight token]{<answer>}} tag.
In the first setting, performance significantly degrades across all metrics, confirming that accurate supervision is critical for learning calibrated confidence.
In the second setting, the model fails to train altogether, suggesting that without an explicit instruction to predict confidence, the model cannot ground the meaning of the scalar and collapses.

\paragraph{Impact of CoT Visibility on the Confidence.}
\label{sec:cot_visibility}

\begin{figure}
\centering
\includegraphics[width=0.7\linewidth]{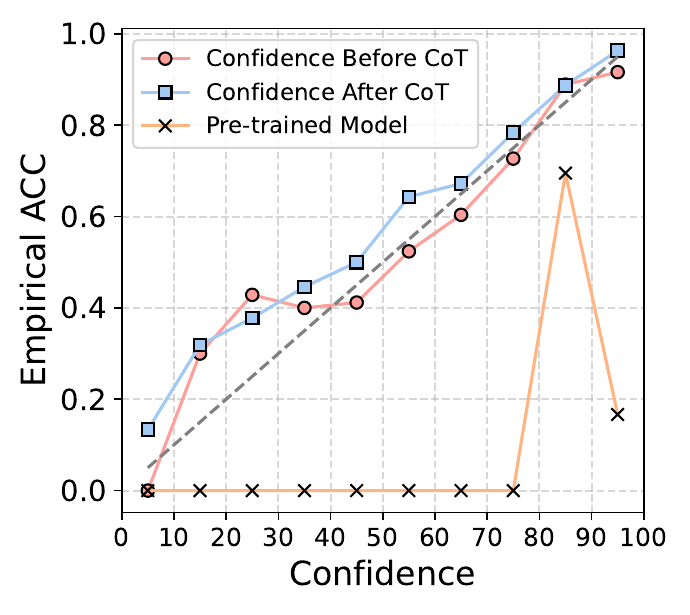}
\vspace{-0.5em}
\caption{Reliability curves of CSFT-trained \texttt{LLaMA3.2-3B-Instruct} on GSM8K, comparing predicted confidence when elicited before vs. after CoT generation. The broadly similar calibration profiles suggest that the model’s confidence reflects internal uncertainty rather than simply CoT length.}
\label{fig:cot_visi}
\end{figure}

To test whether the model's confidence relies on observing the length or content of the generated CoT, we compare calibration when confidence is elicited before versus after CoT generation.
As shown in \Cref{fig:cot_visi}, the two reliability curves are broadly similar, suggesting that the model does not depend heavily on CoT visibility and instead bases its confidence on internal uncertainty.

\subsection{Confidence-Guided Reasoning Path Refinement}

\begin{table}[ht]
\centering
\caption{Manual rethinking improves accuracy in low-confidence bins on GSM8K test set using \texttt{LLaMA3.2-3B-Instruct} (\textbf{$\Delta$ACC}).}
\vspace{0.5em}
\label{tab:delta_acc_rethink}
\renewcommand{\arraystretch}{1.3}
\setlength{\tabcolsep}{6pt}
\small
\begin{tabular}{lcccc}
\toprule
\textbf{Bin} & \textbf{0} & \textbf{10} & \textbf{20} & \textbf{30} \\
\midrule
\textbf{$\Delta$ACC} & +0.5625 & +0.5524 & +0.3158 & +0.2069 \\
\textbf{Count} & 16 & 143 & 38 & 29 \\
\bottomrule
\end{tabular}
\end{table}

When a model is well-calibrated, its verbalized confidence serves as a trustworthy signal for downstream decision-making and control. As demonstrated previously, the CSFT-trained model is capable of accurately predicting its confidence even before generating the full CoT. This capability opens up the possibility of guiding the reasoning trajectory from the very beginning.

We exploit this by manually redirecting low-confidence samples toward alternative reasoning paths. Specifically, if the model expresses low confidence in its initial output(elicited via the prefix prompt shown in \Cref{fig:prefix_prompt}), we initiate a new reasoning attempt with an altered or more structured prompt (as shown in \Cref{fig:low_conf_rethink_prompt}), without waiting for failure. This preemptive rethinking mechanism enables selective refinement with minimal additional cost.

As shown in \Cref{tab:delta_acc_rethink}, this confidence-aware rethinking strategy substantially improves accuracy in the low confidence bins. For example, in the 0–10 confidence range, accuracy improves by over 55 percentage points. This result underscores the utility of confidence not just for post hoc calibration, but also for guiding efficient and cost-aware reasoning-time improvement.
\section{Conclusion and Future Work}

To develop well-calibrated LLMs capable of expressing trustworthy verbalized confidence in CoT reasoning, we propose a simple yet effective fine-tuning method called CSFT. For training, we construct a synthetic dataset based on the GSM8K training set, where each problem is paired with a self-generated confidence label.

Empirically, we demonstrate that LLMs fine-tuned with CSFT achieve substantial improvements across multiple evaluation metrics—including accuracy and ECE—on both the GSM8K test split and out-of-domain benchmarks. Remarkably, we also observe that CSFT elicits the emergence of self-verification behavior, particularly in low-confidence scenarios, despite not providing any explicit supervision related to reasoning strategies during training. Moreover, this self-verification behavior in low-confidence scenarios can act as a valuable mechanism for end users interacting with LLMs. It serves as an implicit indicator that the model recognizes its own uncertainty and is actively attempting to validate its response. 




Looking forward, several directions emerge.
First, since we observe that confidence can be elicited \textit{prior} to reasoning, it may be possible to predict the downstream cost of a reasoning trajectory (e.g., output length or compute usage) from the initial confidence.
This opens up opportunities for confidence-conditioned inference policies that balance accuracy and efficiency. Second, while self-verification is desirable under uncertainty, we find that some low-confidence generations enter excessively long or redundant reasoning loops, potentially reflecting local minima in the generation dynamics. Third, one could explore confidence-aware \textit{steering} of CoT trajectories, or use latent confidence signals to trigger rethink-style interventions without incurring full-generation overhead.


\printbibliography

\newpage
\clearpage
\appendix
\onecolumn
\section{Experimental details}
\label{app:experimental_details}

In this section, we provide detailed information on the models and datasets used in our experiments, along with formal definitions of the calibration metrics employed for evaluation. Specifically, we describe the two instruction-tuned language models used: \texttt{LLaMA3.2-3B-Instruct} and \texttt{Qwen2.5-1.5B-Instruct}. For datasets, we include:
\begin{itemize}
    \item \textbf{GSM8K}~\citep{cobbe2021training}; a dataset of 7.47k grade-school math word problems designed to test step-by-step reasoning. We use 10\% and 20\% of the original training set (0.75k and 1.49k examples, respectively) for training and validation, and the full test set of 1.32k examples for evaluation. Available at \href{https://huggingface.co/datasets/openai/gsm8k}{huggingface.co/openai/gsm8k}.
    
    \item \textbf{MATH-500}~\citep{lightman2023let}; a subset of the MATH dataset consisting of 500 diverse high school level problems covering algebra, geometry, calculus, and more. Used solely for evaluation. Available at \href{https://huggingface.co/datasets/HuggingFaceH4/MATH-500}{huggingface.co/HuggingFaceH4/MATH-500}.
    
    \item \textbf{ARC-Challenge}~\citep{clark2018think}; a multiple-choice science and commonsense QA benchmark containing 1.17k test questions that require reasoning beyond surface-level cues. We use the test set for evaluation. Available at \href{https://huggingface.co/datasets/allenai/ai2_arc}{huggingface.co/allenai/ai2\_arc}.
\end{itemize}

\subsection{Model and datasets}

\begin{table}[ht]
\centering
\caption{Training hyperparameters used for CSFT fine-tuning across both models.}
\label{tab:training_hparams}
\renewcommand{\arraystretch}{1.2}
\setlength{\tabcolsep}{7pt}
\small
\begin{tabular}{lc}
\toprule
\textbf{Hyperparameter} & \textbf{Value} \\
\midrule
Batch size                       & 1 \\
Gradient accumulation            & 16 \\
Learning rate                    & [1e-5, 1e-04] \\
Optimizer                        & AdamW \\
Weight decay                     & 0.0 \\
Warmup ratio                     & 0.0 \\
Max sequence length              & 1024 \\
KL regularization ($\lambda$)    & 0.0 \\
Training steps                   & 2500 \\
Checkpoint selection             & Best dev loss \\
\midrule
\multicolumn{2}{c}{\textbf{LoRA configuration}} \\
LoRA rank ($r$)                  & 128 \\
LoRA alpha                       & 32 \\
LoRA dropout                     & 0.1 \\
LoRA target modules              & $q_\text{proj}$, $v_\text{proj}$ \\
\bottomrule
\end{tabular}
\end{table}

\subsubsection{Calibration metrics}
\label{app:metric_explain}
\begin{itemize}
    \item \textbf{Expected Calibration Error}~\citep[ECE;][]{ece}: 
    \begin{equation*}
    \text{ECE} = \sum_{m=1}^M \frac{|B_m|}{n} \left| \text{acc}(B_m) - \text{conf}(B_m) \right|
    \end{equation*}
    where \( B_m \) is the set of predictions in bin \( m \), \( \text{acc}(B_m) \) is the accuracy, and \( \text{conf}(B_m) \) is the average confidence of the predictions in that bin. ECE measures how well the model’s predicted probabilities are calibrated.

    \item \textbf{Brier Score}~\citep[BS;][]{brier}:
    \begin{equation*}
    \text{BS} = \frac{1}{N} \sum_{i=1}^N (f_i - y_i)^2
    \end{equation*}
    where \( f_i \) is the predicted probability and \( y_i \) is the true label. BS combines both the accuracy and confidence of the predictions, penalizing overconfident and underconfident predictions.

\end{itemize}
\section{Additional Results}

\subsection{Length Analysis on Held-out CoT Tasks}
\label{app:length}

\begin{figure}[t]
\centering
\includegraphics[width=0.5\linewidth]{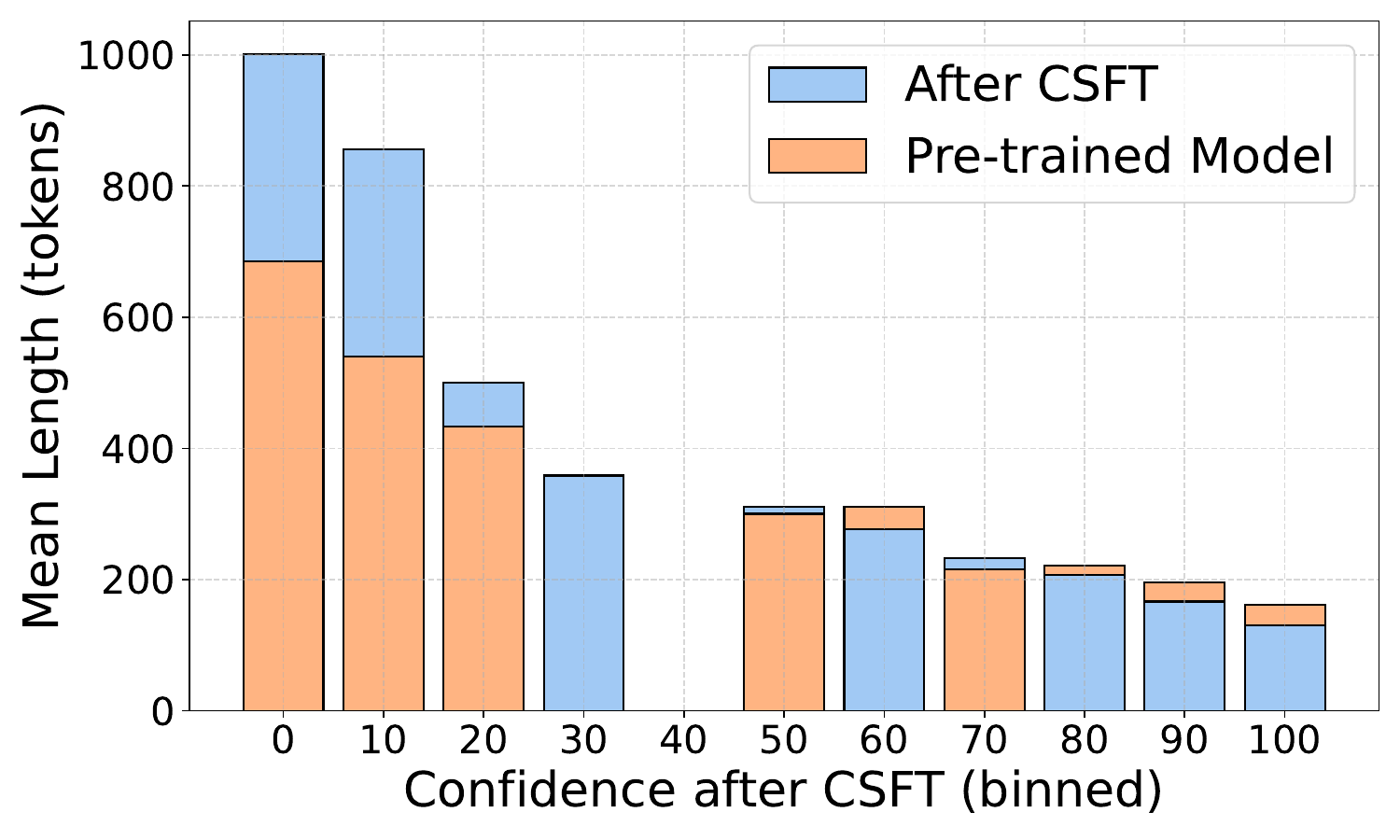}
\vspace{-0.5em}
\caption{Output length across confidence bins on Math-500, using \texttt{LLaMA3.2-3B-Instruct} fine-tuned with CSFT. The model generates significantly longer responses when confidence is low, while high-confidence predictions tend to be more concise.}
\label{fig:math500_length}
\end{figure}

\begin{figure}[t]
\centering
\includegraphics[width=0.5\linewidth]{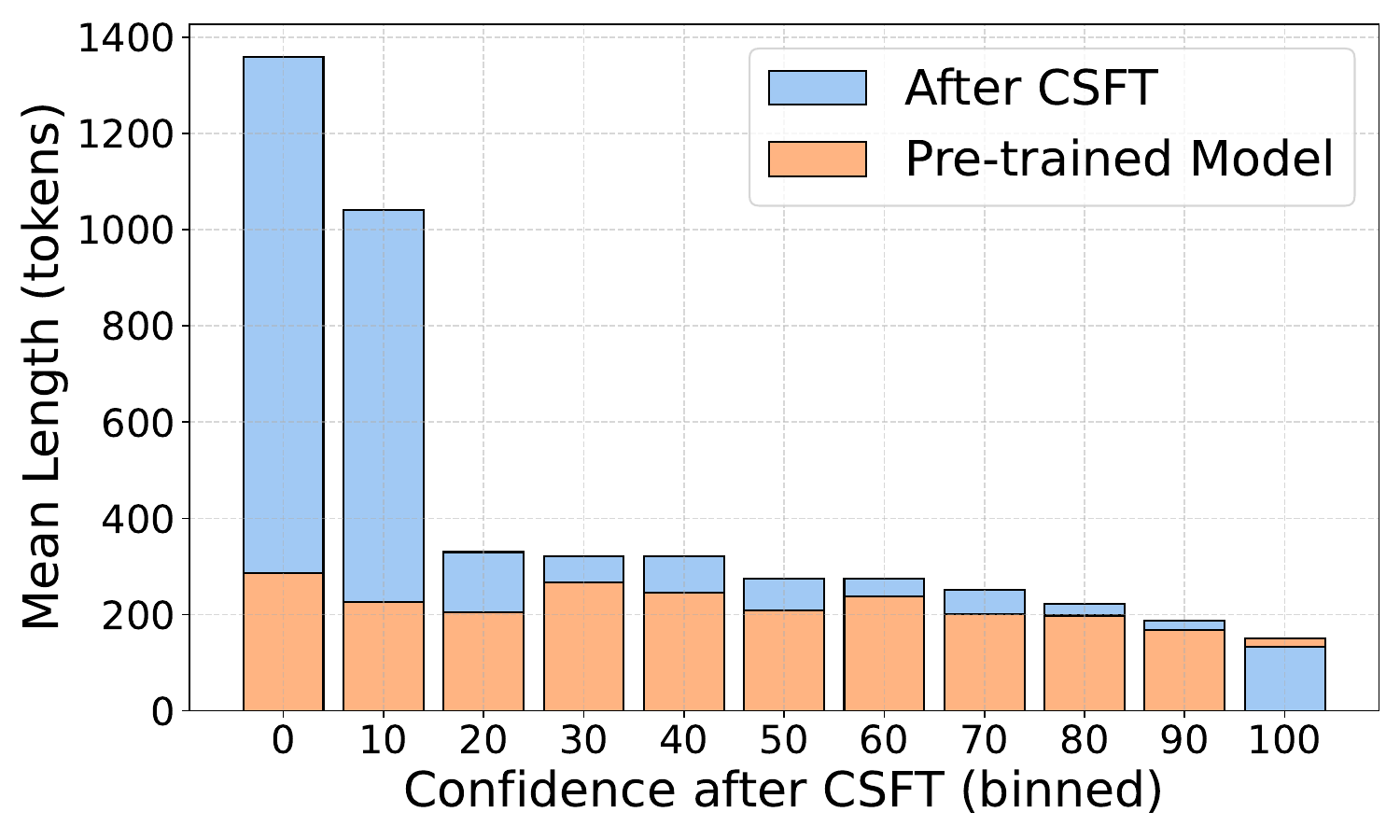}
\vspace{-0.5em}
\caption{Output length across confidence bins on ARC-Challenge, using \texttt{LLaMA3.2-3B-Instruct} fine-tuned with CSFT. The same trend of confidence-conditioned length modulation emerges, despite the task being unseen during training.}
\label{fig:arc_length}
\end{figure}

\Cref{fig:math500_length} and \Cref{fig:arc_length} present an analysis of model outputs on Math-500 and ARC-Challenge—two held-out CoT tasks not seen during CSFT training. As shown in both figures, output length increases in low-confidence bins. In the case of Math-500, there is also a clear trend toward more concise responses in high-confidence bins. These results demonstrate that the length modulation effect reported in the main paper is not restricted to the training distribution but generalizes to unseen tasks. In other words, CSFT enables the model to internalize the ability to adjust response length based on its uncertainty, suggesting a deeper transformation in its reasoning behavior.

\section{Prompt Examples}
\label{app:prompts}

\begin{figure*}[ht]
\centering
\begin{prompt}{Base Reasoning Prompt}
This is a conversation between \textbf{User} and \textbf{Assistant}.

The User asks a question, and the Assistant provides a solution.

Before answering, the Assistant reasons through the problem step-by-step.

The reasoning is enclosed within \verb|<think> ... </think>|,  
and the final answer within \verb|<answer> ... </answer>|.

\vspace{0.5em}
\textbf{Example:}

\verb|{question}|

\verb|<think>{step-by-step reasoning}</think>|

\verb|<answer>{final answer}</answer>|

\vspace{0.5em}
Now, respond to the following using the \textbf{exact same format}:

\verb|<question>|

\end{prompt}
\vspace{-0.5em}
\caption{Base prompt format used to elicit step-by-step reasoning and final answer generation.}
\label{fig:base_prompt}
\end{figure*}

\begin{figure*}[ht]
\centering
\begin{prompt}{Suffix Confidence Prompt}
Please respond with a score from 0 to 100 in
\verb|<confidence> </confidence>| tags.

How confident are you in your previous answer?
\end{prompt}
\vspace{-0.5em}
\caption{Suffix confidence prompt used to elicit model confidence \textit{after} the full reasoning and answer generation. Since confidence is predicted post hoc, this setup allows for calibration without interfering with the model’s original reasoning or final answer.}
\label{fig:suffix_prompt}
\end{figure*}

\begin{figure*}[ht]
\centering
\begin{prompt}{Prefix Confidence Prompt}
This is a conversation between \textbf{User} and \textbf{Assistant}.

The User asks a question, and the Assistant provides a thoughtful, reasoned answer.

Before answering, the Assistant first reasons through the problem step-by-step.

The reasoning is enclosed in \verb|<think> ... </think>| tags.

The final answer is enclosed in \verb|<answer> ... </answer>| tags.

A confidence score is then provided in \verb|<confidence> ... </confidence>| tags,  
representing the Assistant’s certainty as a \textbf{continuous value between 0 and 100}.

\vspace{0.5em}
\textbf{Example:}

\verb|{question}|

\verb|<think>{step-by-step reasoning}</think>|

\verb|<answer>{final answer}</answer>|

\verb|<confidence>{confidence}</confidence>|

\vspace{0.5em}
Now, answer the following in \textbf{exactly} the same format:

\verb|<question>|

\end{prompt}
\vspace{-0.5em}
\caption{Prefix confidence prompt used to elicit model confidence \textit{before} reasoning begins. By conditioning the generation on anticipated confidence, this prompt not only guides the model's uncertainty expression but also influences the reasoning path and final answer.}
\label{fig:prefix_prompt}
\end{figure*}

\begin{figure*}[ht]
\centering
\begin{prompt}{GSM8K Parsing Prompt}
\textbf{Instruction:}

We have a user's question and a model's generated response:

\textbf{Your task:}\\  
1. Carefully read the question and the generated response in \textbf{Example 6 only}.\\  
2. Extract the final answer based on the following rules:
\begin{itemize}
  \item If the response contains a number (with or without units), \textbf{extract only the numeric value}.
  \item If the response is purely textual (no numbers), \textbf{extract the exact string as it appears}.
\end{itemize}
3. Use the following output format:  
\quad \texttt{Model's Final Answer is: [Your extracted answer]}

\vspace{0.5em}
\textbf{Rules:}
\begin{itemize}
  \item Only process \textbf{Example 6} for extraction. Ignore all other examples.
  \item Do not include units, symbols, or extra text when extracting numbers.
  \item Provide the answer strictly in the requested format without additional explanations.
\end{itemize}

\vspace{0.5em}
\textbf{Examples}

\textbf{Example 1:}  
Model's Generated Response: It takes about 160 minutes.\\  
\texttt{Model's Final Answer is: 160}

\textbf{Example 2:}  
Model's Generated Response: The nearest star is approximately 4.24 light years away.\\  
\texttt{Model's Final Answer is: 4.24}

\textbf{Example 3:}  
Model's Generated Response: The tallest mountain is Mount Everest.\\  
\texttt{Model's Final Answer is: Mount Everest}

\textbf{Example 4:}  
Model's Generated Response: It weighs 5 kg.\\  
\texttt{Model's Final Answer is: 5}

\textbf{Example 5:}  
Model's Generated Response: 81 + 221 - 24 = 278.\\  
\texttt{Model's Final Answer is: 278}

\vspace{0.5em}
\textbf{Example 6:}  
Model's Generated Response: \texttt{<answer\_text>}

\end{prompt}
\vspace{-0.5em}
\caption{Prompt used to extract final answers from model-generated responses on GSM8K.}
\label{fig:gsm8k_parsing_prompt}
\end{figure*}

\begin{figure*}[ht]
\centering
\begin{prompt}{Math500 Matching Prompt}
\textbf{Instruction:}

You are given the true answer and the final answer generated by a model for a math problem.

\textbf{Your task:}
\begin{enumerate}
  \item Only examine \textbf{Example 6}.
  \item Compare the \textbf{model's final answer} and the \textbf{true answer}.
  \item Respond with \texttt{"yes"} if they exactly match, otherwise respond with \texttt{"no"}.
  \item Do not include any explanation or extra words—just respond with \texttt{"yes"} or \texttt{"no"}.
\end{enumerate}

\vspace{0.5em}
\textbf{Examples}

\textbf{Example 1:} \\
True Answer: 0.5 \\
Model Answer: 1/2 \\
Is it correct?: \texttt{yes}

\textbf{Example 2:} \\
True Answer: 24 \\
Model Answer: 22 \\
Is it correct?: \texttt{no}

\textbf{Example 3:} \\
True Answer: 8 \\
Model Answer: 32 / 4 = 8 \\
Is it correct?: \texttt{yes}

\textbf{Example 4:} \\
True Answer: \texttt{\textbackslash frac\{10\}\{4\}} \\
Model Answer: \texttt{\textbackslash frac\{9\}\{4\}} \\
Is it correct?: \texttt{no}

\textbf{Example 5:} \\
True Answer: 3 \\
Model Answer: \texttt{\textbackslash frac\{15\}\{5\}} \\
Is it correct?: \texttt{yes}

\vspace{0.5em}
\textbf{Example 6:} \\
True Answer: \texttt{<true\_answer>} \\
Model Answer: \texttt{<model\_answer>} \\
Is it correct?:
\end{prompt}
\vspace{-0.5em}
\caption{Matching prompt for evaluating exact answer agreement on Math500. Designed to assess correctness by comparing model output with the ground truth in a strict yet interpretable format.}
\label{fig:math500_matching_prompt}
\end{figure*}

\begin{figure*}[ht]
\centering
\begin{prompt}{ARC-Challenge Parsing Prompt}
\textbf{Instruction:}

A user's question provides four choices formatted exactly as:
\begin{verbatim}
A. <option A> 
B. <option B>
C. <option C>
D. <option D>
\end{verbatim}

We also have the model's generated response.

\textbf{Your task:}
\begin{enumerate}
  \item Read \textbf{Example 6 only}.
  \item Decide which single choice (\texttt{A}, \texttt{B}, \texttt{C}, or \texttt{D}) the model ultimately selected, following these rules:
  \begin{itemize}
    \item \textbf{Letter match} – If the response explicitly includes the letter ‘A’, ‘B’, ‘C’, or ‘D’ (optionally followed by punctuation), extract \textbf{only that letter}.
    \item \textbf{Text match} – If no letter is given, compare the response text (case-insensitive, ignoring punctuation and surrounding spaces) with each option; if it matches exactly one, return the corresponding letter.
    \item If both a letter and option text appear, treat the letter as authoritative.
  \end{itemize}
  \item Output format (strict): \\
  \texttt{Model's Final Answer is: <A | B | C | D>}
\end{enumerate}

\textbf{Do not add explanations or any extra text.}

\vspace{0.5em}
\textbf{Examples}

\textbf{Example 1:} \\
Choices:  
A. Paris \quad B. Berlin \quad C. Madrid \quad D. Rome \\
Model's Generated Response: A. Paris is the capital of France. \\
\texttt{Model's Final Answer is: A}

\textbf{Example 2:} \\
Choices:  
A. 3 \quad B. 4 \quad C. 5 \quad D. 6 \\
Model's Generated Response: The correct option is B. \\
\texttt{Model's Final Answer is: B}

\textbf{Example 3:} \\
Choices:  
A. Spring \quad B. Summer \quad C. Autumn \quad D. Winter \\
Model's Generated Response: It’s usually coldest in winter. \\
\texttt{Model's Final Answer is: D}

\vspace{0.5em}
\textbf{Example 6:} \\
Choices: \texttt{\{choices\}} \\
Model's Generated Response: \texttt{\{answer\_text\}} \\
\end{prompt}
\vspace{-0.5em}
\caption{Parsing prompt for multiple-choice answer extraction on ARC-Challenge. The rules prioritize explicit letter selection, with fallback to semantic string matching.}
\label{fig:arc_parsing_prompt}
\end{figure*}

\begin{figure*}[ht]
\centering
\begin{prompt}{Pre-CoT Confidence Prompt}
Before generating your answer, can you first assess your internal confidence (0--100) in its correctness and state it using `<confidence> </confidence>` tags, then proceed to provide your full answer?
\end{prompt}
\vspace{-0.5em}
\caption{Prompt format for eliciting a model’s self-assessed confidence \textbf{prior to} generating CoT response.}
\label{fig:pre_cot_conf_prompt}
\end{figure*}

\begin{figure*}[ht]
\centering
\begin{prompt}{Low Confidence Rethinking Prompt}
Your confidence score is low. Rather than following your current reasoning path, pause and explore an alternative approach that is likely to raise your confidence.
Think step-by-step and provide a revised answer.
\end{prompt}
\vspace{-0.5em}
\caption{Prompt used when the model reports low confidence, encouraging it to pause and reconsider its reasoning path to generate a more confident response.}
\label{fig:low_conf_rethink_prompt}
\end{figure*}

This section presents the prompt templates used in our experiments for eliciting model reasoning, answers, and confidence scores. 
We include both the \textit{prefix} prompt, where the confidence is generated before reasoning begins, and the \textit{suffix} prompt, where confidence is predicted after the final answer. 
All prompts follow a standardized format to ensure consistent supervision during CSFT and reliable evaluation during inference.

\end{document}